\documentclass{article} 
\usepackage{iclr2027_conference,times}
\setcitestyle{square,numbers,sort&compress,citesep={,}}
\usepackage[pagebackref=true,breaklinks=true,letterpaper=true,colorlinks,bookmarks=false]{hyperref}
\usepackage{tcolorbox}
\tcbuselibrary{skins}
\usepackage{enumitem}
\usepackage{amsmath}
\usepackage{amssymb}

\newcommand{\ie}{\textit{i}.\textit{e}.}

\newcommand{\cf}{\textit{cf}.}

\usepackage{amsmath,amsfonts,bm}

\def\eqref#1{equation~\ref{#1}}

\def\1{\bm{1}}

\DeclareMathAlphabet{\mathsfit}{\encodingdefault}{\sfdefault}{m}{sl}
\SetMathAlphabet{\mathsfit}{bold}{\encodingdefault}{\sfdefault}{bx}{n}

\usepackage{hyperref}
\usepackage{url}

\usepackage{fontawesome}
\usepackage{libertine}

\usepackage{xspace}
\usepackage{pifont}
\usepackage{caption}
\usepackage{wrapfig}

\usepackage{booktabs}
\usepackage{multirow}
\usepackage{graphicx}
\usepackage[table]{xcolor}
\usepackage{arydshln} 
\usepackage{subcaption}

\definecolor{oursblue}{RGB}{230,240,251}
\definecolor{groupgray}{RGB}{246,243,241}
\definecolor{gaincolor}{RGB}{38,148,166}
\definecolor{mygray}{gray}{.9}
\makeatletter
\newcommand{\thickhline}{%
    \noalign {\ifnum 0=`}\fi \hrule height 1pt
    \futurelet \reserved@a \@xhline
}

\newcommand{\ourmethod}{{\fontfamily{lmtt}\selectfont \textbf{LatentStream}}\xspace}

\title{Beyond Retrieval: Progressive Latent Memory\\ Evolution for Streaming Video Understanding}

\author{Hongyu Qu$^{*1}$, Guangming Yao$^{* \dagger 2}$, Ling Xing$^{1}$, Xiaobin Hu$^{3}$, Rongxing Ding$^{1}$, \\ \textbf{Guibin Zhang$^{3}$}, \textbf{Fan Zhang$^{4}$}, \textbf{Yi Yuan$^{2}$}, \textbf{Xiangbo Shu$^{\dagger1}$}, \textbf{Shuicheng Yan$^{3}$}\\
  {\small $^1$Nanjing University of Science and Technology \quad $^2$Ant Group\quad $^3$ National University of Singapore}\\ 
  {\small $^4$ The Chinese University of Hong Kong \quad $*$ Equal contribution \quad $\dagger$ Corresponding authors }\vspace{.5em} \\
}

\iclrfinalcopy 
\begin{document}

\maketitle

\begin{abstract}
Streaming video understanding requires multimodal large language models (MLLMs) to process continuous visual inputs and respond
to user queries under strict causality and bounded memory.
Existing approaches typically compress historical observations into an external memory bank and retrieve query-relevant evidence as additional visual context.
Though effective, this store-and-retrieve paradigm keeps historical evidence as external visual context, preventing it from being internalized into a compact, evolving latent memory that can continuously guide streaming reasoning.
To bridge this gap, we introduce \ourmethod, a progressive latent working memory framework that shifts streaming memory from \textit{store-and-retrieve} to \textit{retrieve-and-internalize}.
Specifically, \ourmethod comprises three coordinated components. First, Query-Agnostic Hierarchical Streaming Memory organizes visual history into short-, mid-, and long-term levels under a fixed memory budget through Jenks-guided adaptive consolidation. Once a query arrives, Hierarchical Latent Memory Evolution equips groups of latent memory tokens with progressively expanding memory receptive fields, enabling them to iteratively retrieve historical evidence from their corresponding scopes and internalize it into a compact, fixed-length latent memory. Finally, Progressive Confidence-guided Latent Memory Optimization constructs a hierarchical progression reward from group-wise predictive entropy and jointly refines the latent memory tokens and retrieved evidence, encouraging increasingly confident streaming reasoning.
Extensive experiments demonstrate that \ourmethod achieves new state-of-the-art results on existing online and offline video benchmarks. 
\end{abstract}

\section{Introduction}
\begin{wrapfigure}[17]{r}{0.57\textwidth}
\vspace{-.5cm}
        \hspace{+0.04cm}
        \centering
		\includegraphics[width=0.99\linewidth]{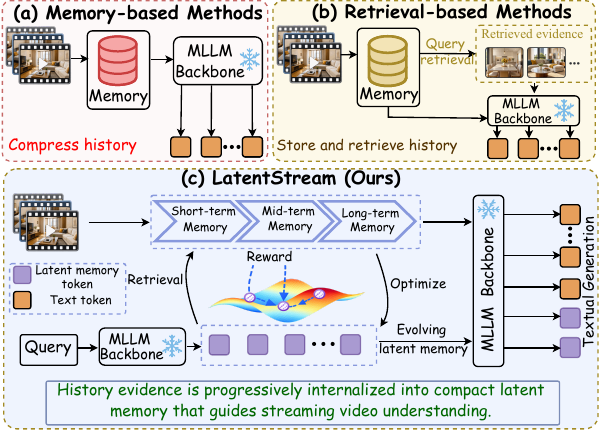}
\captionsetup{font=small,width=1\linewidth}
 \vspace{-0.3cm}
	\caption{\small{Motivation of our \ourmethod: paradigm comparison of memory-based and retrieval-based methods with our retrieve-and-internalize method \ourmethod. 
 }}
 \label{fig:motivation}
\end{wrapfigure}
Streaming video understanding requires models to continuously process incoming visual observations and respond to user questions that may be posed at any time \citep{di2025streaming,xiong2025streaming,xu2026streamingvlm}. This capability is essential for a wide range of real-world applications, including live monitoring~\cite{dumitru2026multi}, autonomous driving~\cite{chen2024end,brodermann2025cafuser}, smart glasses~\cite{zhang2026egonight,lee2018interaction}, as well as embodied agents and robotic systems~\cite{wang2026ocra,liu2025aligning}.

To equip existing Video-LLMs~\cite{zhang2024llava,qwen25,Llavaonevision,wang2024qwen2} for continuously arriving visual streams, recent efforts have explored streaming video understanding from multiple perspectives, ranging from dedicated model training~\cite{liu2026thinking,guan2026video,zhang2026think} to streaming inference~\cite{xu2026streamingvlm} and memory-based context management~\cite{fluxmem,oasis,wu2026semantic}. 
Among them, \textit{training-free} approaches are more appealing as they directly adapt off-the-shelf Video-LLMs to streaming scenarios without additional parameter updates (Fig.~\ref{fig:motivation}a).
Their key strategy is to manage the continuously accumulated visual context at inference time through frame selection~\cite{shen2026simple}, visual token pruning or merging~\cite{dorovatas2026recurrent,timechat}, KV-cache management~\cite{pang2026decouple,chen2026streamkv}, and fixed-capacity or hierarchical memory~\cite{fluxmem}.
More recent approaches~\cite{oasis,wu2026semantic} further couple such memory management with query-aware retrieval, selectively recalling relevant evidence from the retained history once a query arrives (Fig.~\ref{fig:motivation}b).
  By filtering redundant observations and retaining potentially informative evidence, these methods effectively constrain the growth of historical context, reducing memory and computational overhead for long-form video streams.

Despite these advances, existing methods typically treat streaming memory as an external bank of historical evidence, focusing primarily on what information to retain and what to retrieve once a query arrives~\cite{oasis,fluxmem,wu2026semantic}.
 Although query-aware memory retrieval allows the model to identify relevant visual evidence from the retained history,  the retrieved evidence is still exposed to the model as external, variable-length visual context.
Such a retrieval-centric paradigm essentially addresses which historical evidence should be accessed, while leaving largely unexplored how the accessed evidence can be internalized into a compact latent state that participates directly in subsequent video reasoning. 
As a result, query-agnostic streaming memory and query-conditioned reasoning remain loosely coupled: \textit{there is no compact latent memory representation that can accumulate task-relevant historical evidence and continuously evolve alongside the video reasoning process in the model latent space}. 
This work posits that the model's latent space~\cite{liu2026reasoning,chen2025reasoning,wang2026monet,yang2026machine} provides a natural substrate for bridging this gap.

Building on this view, we argue that streaming video reasoning calls for a \textbf{latent working memory}: 
rather than exposing retrieved visual evidence to the model as variable-length reasoning context, such a latent memory progressively internalizes task-relevant history into a fixed-length latent state that further guides the streaming video reasoning process. 
Crucially, this latent state is not a one-shot compression of the retrieved history, but continuously evolves with newly accessed evidence and, in turn, further guides the retrieval of relevant historical information.
This creates an iterative interplay between external memory access and latent memory evolution, turning the conventional store-and-retrieve mechanism into a retrieve-and-internalize paradigm. 
This naturally raises our pivotal research question: 
\begin{tcolorbox}[
    enhanced,
    sidebyside, 
   colframe=black!70,
        colback=cyan!4,
    boxrule=1pt, arc=4mm,
    lefthand width=0.06\linewidth,
    sidebyside gap=5mm]
\includegraphics[width=\linewidth]{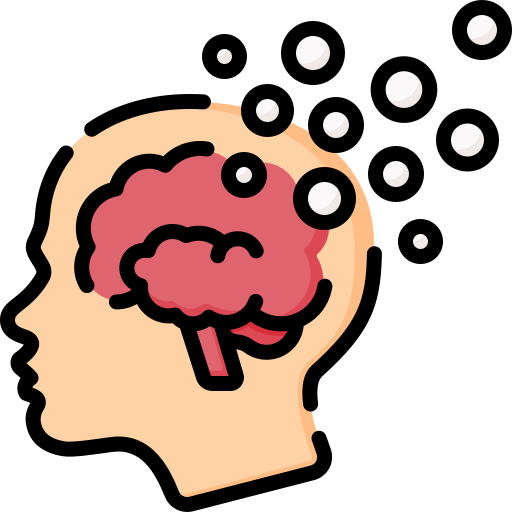}

\tcblower 

\textit{How can historical evidence in external streaming memory be adaptively internalized into a compact, query-conditioned, and continuously evolving latent memory for streaming video reasoning?}
\end{tcolorbox}
To answer this question, we introduce \ourmethod, a progressive latent working memory framework for streaming Video-LLMs (Fig.~\ref{fig:motivation}c).
Instead of treating retrieved history as auxiliary context, \ourmethod progressively internalizes task-relevant visual evidence into a compact, evolving latent memory that continuously guides streaming reasoning.
At its core, \ourmethod comprises three coordinated components.
First, \ourmethod constructs a \ding{168} \textbf{Query-agnostic Hierarchical Streaming Memory (HSM)} that organizes incoming observations into short-, mid-, and long-term memories under a fixed budget.
A Jenks-guided adaptive consolidation strategy progressively reduces temporal and spatial redundancy while preserving informative evidence across extended video streams.
Second, \ourmethod introduces \ding{169} \textbf{Hierarchical Latent Memory Evolution (HME)} to internalize such hierarchical streaming memory into Latent Memory Tokens (LMTs), which are divided into three groups with progressively expanding memory receptive fields.
At each evolution iteration, the LMTs retrieve relevant evidence from their respective memory scopes and evolve jointly with the retrieved visual memory, subsequently guiding the next round of historical access, forming iterative memory retrieval--update evolution.
Third, to regulate this latent memory evolution, \ourmethod develops \ding{170} \textbf{Progressive Confidence-guided Latent Memory Optimization (PMO)}, which constructs a hierarchical confidence progression reward from group-wise latent token predictive entropy that encourages increasingly confident reasoning as the accessible historical scope expands. Guided by this objective, the latent memory tokens progressively absorb task-relevant evidence through test-time optimization without modifying the Video-LLM parameters.
In this way, historical memory is no longer merely appended as variable-length auxiliary context, but is selectively internalized into a compact latent working memory that directly participates in subsequent reasoning.

We extensively validate \ourmethod on five video understanding benchmarks spanning both online and offline settings. The results show that \ourmethod achieves new state-of-the-art or highly competitive performance across diverse tasks. Specifically, it achieves \textbf{64.2\%} on OVO-Bench~\cite{ovobench} and \textbf{76.9\%} on StreamingBench~\cite{streamingbench} for streaming evaluation, while attaining \textbf{66.6\%} on VideoMME~\cite{videomme}, \textbf{74.0\%} on MLVU~\cite{mlvu}, and \textbf{62.1\%} on LongVideoBench~\cite{longvideobench} for offline long-video understanding. These results demonstrate that a latent working memory framework can consistently improve both online and offline video understanding under a bounded memory budget.

In summary, our main contributions are as follows:
\begin{itemize}[leftmargin=*]
\item We introduce \ourmethod, a progressive latent working memory framework that advances conventional \textit{store-and-retrieve} memory toward a \textit{retrieve-and-internalize} paradigm, progressively internalizing task-relevant historical evidence into compact and evolving latent memory that can continuously guide streaming video reasoning.
\item We propose Hierarchical Latent Memory Evolution, which couples Jenks-guided hierarchical streaming memory with three latent memory token groups of expanding memory receptive fields, enabling them to iteratively retrieve historical evidence from their corresponding scopes and internalize it into a compact, fixed-length latent memory.
\item We develop Progressive Confidence-guided Latent Memory Optimization, which adopts a hierarchical progression reward based on group-wise latent token predictive entropy and refines the latent memory token embeddings, encouraging increasingly confident streaming reasoning.
\end{itemize}

\section{Related Work}
\label{sec:related}
\noindent\textbf{Streaming Video Understanding.}
 Unlike offline video understanding, which assumes the whole video is accessible beforehand, streaming video understanding requires models to causally process continuously arriving observations and respond to user queries in real time. 
 Existing studies can be broadly categorized into four groups.
 \textbf{(i)} \textit{Proactive interaction} methods aim to determine not only what to respond but also when to respond, typically through response prediction heads~\cite{azad2026streamready,yan2026proact}, generative trigger tokens~\cite{xia2026streaming,zhang2025proactive}, event-aware activation mechanisms~\cite{fluxmem,guo2026event}, or reinforcement learning~\cite{liu2026thinking}.
 \textbf{(ii)} \textit{Streaming memory} methods~\cite{fluxmem,wu2026semantic,oasis} maintain useful historical information under bounded context and computation budgets, enabling models to access past observations during interaction.
  \textbf{(iii)} More recently, \textit{Streaming thinking} methods~\cite{zhang2026think,liu2026thinking,guan2026video} further couple perception with reasoning, allowing intermediate reasoning states to evolve progressively with incoming observations instead of postponing reasoning until a query arrives.
  \textbf{(iv)}  In parallel, substantial efforts are devoted to \textit{real-time inference}, reducing the cost of continuous video processing via selective model invocation~\cite{kim2026stride,ding2025streammind}, visual token reduction~\cite{wang2026accelerating,wu2024videollm}, or KV-cache optimization~\cite{zhang2026hermes,kim2026infinipot}. Together, these advances progressively extend video-language models from offline video processing toward causal, persistent, and real-time understanding and interaction in continuously evolving visual environments.

\noindent\textbf{Long-term Memory Management in Streaming Videos.}
 A fundamental challenge in streaming video understanding is to preserve useful historical information from an unbounded visual stream under bounded memory and context budgets. 
 Existing approaches can be broadly grouped into four categories. 
 \textbf{(i)} \textit{Hierarchical multi-level memory}~\cite{wang2026curvestream,xiong2025streaming,fluxmem} organizes historical observations at different temporal scales or granularities, typically preserving detailed recent context while progressively consolidating older observations into compact long-term representations. 
  \textbf{(ii)} \textit{Visual token compression and pruning}~\cite{timechat,li2025videoscan} control memory growth by removing spatially or temporally redundant visual tokens while retaining informative content. 
  \textbf{(iii)} A closely related line develops \textit{KV-cache memory}~\cite{di2025streaming,ning2025livevlm,yang2025streammem,zhang2026hermes}, which directly compresses, retrieves, or reuses cached internal states to bound GPU memory and avoid repeatedly encoding historical observations. 
   \textbf{(iv)} \textit{Retrieval-augmented memory}~\cite{zhao2026cogstream,oasis} instead decouples long-term storage from the active reasoning context, maintaining historical visual features or compressed memories externally and retrieving query-relevant evidence on demand.
   More fundamentally, existing streaming memory methods primarily consume retrieved history as auxiliary visual context or KV cache, rather than maintaining an explicit working state that progressively evolves with historical evidence.
In contrast, \ourmethod internalizes retrieved evidence into compact latent memory tokens that iteratively evolve and guide subsequent memory access, enabling historical information to directly participate in streaming video reasoning.

\begin{figure}[!t]
    \centering
    \includegraphics[width=0.99\textwidth]{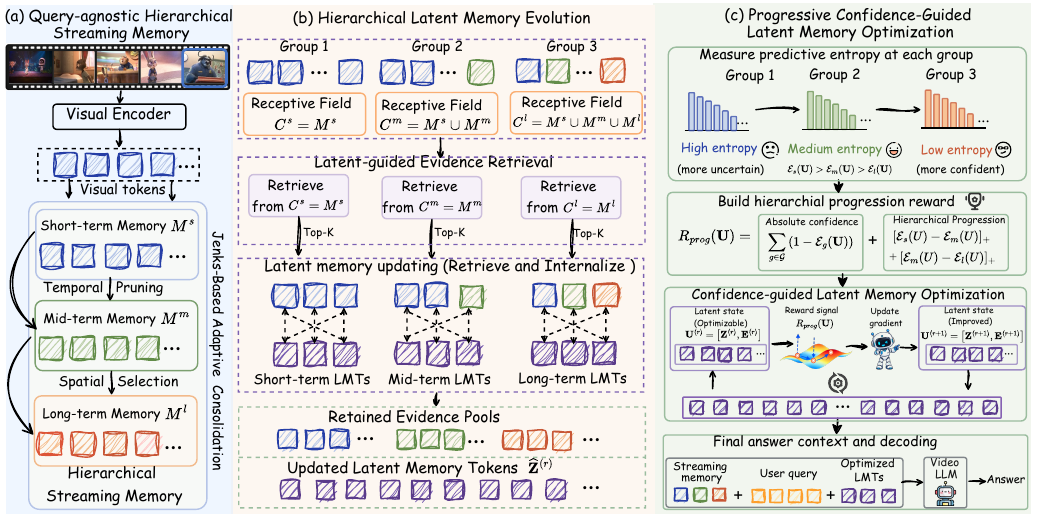}
    \vspace{-10pt}
    \caption{The overview of \ourmethod.
    (a) Query-agnostic hierarchical streaming memory (HSM) organizes incoming visual tokens into short-, mid-, and long-term memories via  Jenks-based adaptive consolidation. (b)  Hierarchical latent memory evolution (HME) retrieves and internalizes historical evidence into compact latent memory via progressively expanding memory receptive fields. (c) Progressive confidence-guided latent memory optimization (PMO) refines the latent memory tokens with a hierarchical progression reward for streaming reasoning.}
    \label{fig:overview}
    \vspace{-15pt}
\end{figure}

\definecolor{revisiongreen}{RGB}{0,135,70}
\section{Methodology}
\label{sec::method}

\subsection{Framework Overview}
\label{sec:framework_overview}
We propose \ourmethod (Fig.~\ref{fig:overview}), a progressive latent working memory framework for streaming video understanding. First, \textbf{Query-agnostic Hierarchical Streaming Memory} continuously organizes incoming visual tokens into short-, mid-, and long-term memories under a fixed token budget. While recent observations are densely retained, a Jenks-guided hierarchical routing mechanism performs tri-level temporal routing over mid-term visual evidence and aggressive spatial consolidation over long-term visual evidence.
Second, \textbf{Hierarchical Latent Memory Evolution} equips different groups of Latent Memory Tokens (LMTs) with progressively expanding memory receptive fields. Through selective latent-guided historical evidence retrieval and injection in the frozen video-LLM, these LMTs progressively absorb critical visual evidence across temporal scales into a compact, query-conditioned latent memory. 
Third, \textbf{Progressive Confidence-guided Latent Memory Optimization} estimates the prediction uncertainty associated with different LMT groups and constructs a hierarchical confidence progression reward, encouraging increasingly confident reasoning as the accessible historical scope expands.
Guided by this objective, the additionally retrieved visual token embeddings and the LMT embeddings are jointly optimized at test time.
After optimization, the retrieved visual tokens are removed from the generation context, and the final answer is generated from the bounded external memory, the query, and the optimized latent memory tokens.
\subsection{Query-agnostic Hierarchical Streaming Memory}
\noindent\textbf{Multi-level Streaming Memory Bank.}
To accommodate an ever-growing video stream within a bounded memory budget, we construct a query-agnostic hierarchical memory \(\mathcal{M}=\{\mathcal{M}^{g}\}_{g\in\mathcal{G}}\), where \(\mathcal{G}=\{s,m,l\}\) indexes the short-, mid-, and long-term levels.
Following the hierarchical consolidation paradigm~\cite{fluxmem}, incoming visual tokens are first stored densely in the short-term memory \(\mathcal{M}^{s}\) to preserve recent perceptual details.
As the stream evolves, historical representations are progressively migrated across memory levels through \textit{Jenks-guided adaptive consolidation}: overflowing short-term evidence is selectively routed into \(\mathcal{M}^{m}\) with different degrees of temporal preservation, while older mid-term evidence is consolidated into \(\mathcal{M}^{l}\) under stronger spatial compression.
Notably, this memory is constructed entirely online before the query arrives, providing a query-agnostic historical basis for the subsequent latent memory reasoning.

\noindent\textbf{Jenks-Based Adaptive Memory Consolidation.}
To adapt memory compression to the continuously changing redundancy patterns of streaming videos, we employ \textit{Jenks Natural Breaks}~\cite{jenks1971error} to derive data-dependent partitions directly from temporal and spatial score distributions.
For short-to-mid memory transition, three-class Jenks partitioning is applied to temporal importance scores, yielding two breakpoints that divide historical representations into \emph{drop}, \emph{compress}, and \emph{preserve} groups.
For the short-to-mid transition, three-class Jenks partitioning produces two breakpoints that route visual representations into \texttt{Drop}, \texttt{Compress}, or \texttt{Preserve}:
\begin{equation}
\rho_{\mathrm{temp}}(s_i)
=
\begin{cases}
\texttt{Drop},      & s_i < \tau_1,\\
\texttt{Compress},  & \tau_1 \leq s_i < \tau_2,\\
\texttt{Preserve},  & s_i \geq \tau_2,
\end{cases}
\quad
(\tau_1,\tau_2)
=
\operatorname{Jenks}_{3}
\left(\mathcal{S}_{\mathrm{temp}}\right),
\label{eq:jenks_memory_routing}
\end{equation}
here \(\mathcal{S}_{\mathrm{temp}}=\{s_i\}\) is the temporal-importance distribution, and \(s_i\) is estimated from the local cosine-distance variation between a visual token and its spatial neighbors across adjacent frames following~\cite{fluxmem}.
Low-score representations are discarded, intermediate ones are temporally compressed, and high-score representations are preserved.
For the mid-to-long transition, we apply two-class Jenks partitioning to the spatial-distance distribution \(\mathcal{S}_{\mathrm{spat}}=\{d_{ij}\}\), where \(d_{ij}=1-\cos(\mathbf{v}_i,\mathbf{v}_j)\) measures the feature distance between neighboring preserved tokens.
Pairs assigned to the low-distance group are spatially redundant and therefore merged, whereas those in the high-distance group are retained individually. More details about hierarchical streaming memory construction are provided in Appendix.

\subsection{Hierarchical Latent Memory Evolution}
Although the hierarchical memory bank maintains a bounded summary of the streaming history, it remains an external and query-agnostic repository. Exposing this memory to the model provides historical context,  but it does not yield a compact latent state that can internalize task-relevant evidence to guide streaming video reasoning. To bridge external memory retention and query-conditioned reasoning, we introduce a set of hierarchical Latent Memory Tokens (LMTs) that repeatedly retrieve historical evidence, interact with the retrieved visual tokens, and evolve in the latent space of a frozen MLLM.

\noindent\textbf{Grouped LMTs with Expanding Memory Receptive Fields.}
Given the memory bank \(\mathcal{M}=\{\mathcal{M}^{g}\}_{g\in\mathcal{G}}\), we partition the latent memory tokens at evolution iteration \(r\) into three groups:
\begin{equation}
    \widehat{\mathbf{Z}}^{(r)}
    =
    \left[
        \widehat{\mathbf{Z}}_{s}^{(r)};
        \widehat{\mathbf{Z}}_{m}^{(r)};
        \widehat{\mathbf{Z}}_{l}^{(r)}
    \right],
    \qquad
    \widehat{\mathbf{Z}}_{g}^{(r)}
    \in
    \mathbb{R}^{K \times D},
    \quad
    g \in \{s,m,l\},
\end{equation}
where \(K\) denotes the number of LMTs in each group, and
\(D\) is the embedding dimension of the MLLM.
To align the LMT hierarchy with the temporal organization of the external memory, we assign the three groups nested retrieval receptive fields:
\begin{equation}
\begin{aligned}
    \mathcal{C}^{s} = \mathcal{M}^{s}, \qquad
    \mathcal{C}^{m}= \mathcal{M}^{s}\cup \mathcal{M}^{m}, \qquad
    \mathcal{C}^{l}= \mathcal{M}^{s} \cup \mathcal{M}^{m}\cup \mathcal{M}^{l}.
\end{aligned}
\label{eq:expanding_memory_receptive_fields}
\end{equation}
Thus, the three groups progressively extend their accessible
history from recent observations to the complete memory bank.
This cumulative design allows broader-range LMTs to integrate
long-term evidence without losing access to recent finer information.
A bootstrap forward pass contextualizes the initial LMT embeddings \(\mathbf{Z}^{\mathrm{init}}\) jointly with the query and the fixed external memory under a group-specific attention mask, producing the
query-conditioned initialization \(\widehat{\mathbf{Z}}^{(0)}\) for the
subsequent retrieval--update memory evolution.

\noindent\textbf{Latent-guided Historical Evidence Retrieval.}
As the latent memory progressively internalizes historical information, the evidence required for its subsequent evolution may change accordingly.  To this end, we dynamically refresh the retrieved context at each evolution iteration. Specifically, at iteration \(r\), each LMT group \(\widehat{\mathbf{Z}}_{g}^{(r-1)}\) retrieves historical evidence exclusively from its corresponding memory receptive field \(\mathcal{C}^{g}\). For each candidate visual token \(\mathbf{m}_{j} \in \mathcal{C}^{g}\), we define \(a_{g,j}^{(r)}\) as its maximum cosine similarity to the \(K\) LMTs in \(\widehat{\mathbf{Z}}_{g}^{(r-1)}\). This maximum cosine similarity preserves a strong affinity between the candidate evidence and any
individual LMT. The retrieved evidence is then given by:
\begin{equation}
\begin{aligned}
\mathcal{I}_{g}^{(r)}=
\operatorname{TopK}_{B}
\left(
\left(
a_{g,j}^{(r)}
\right)_{\mathbf{m}_{j}\in\mathcal{C}_{g}}
\right),  \qquad
\mathbf{E}_{g}^{(r)}=
\left[
\mathbf{m}_{j}
\right]_{j\in\mathcal{I}_{g}^{(r)}},
\end{aligned}
\label{eq:latent_guided_retrieval}
\end{equation}
where $\operatorname{TopK}_{B}(\cdot)$ returns the indices of the $B$ highest-scoring candidate visual evidence, where the same retrieval budget $B$ is used for all LMT groups, and $\mathbf{E}_{g}^{(r)}$ denotes the temporary visual evidence retrieved for group $g$. Although $q$ is not directly employed as a retrieval vector, its semantics have already been encoded into $\widehat{\mathbf{Z}}^{(0)}$ through the bootstrap forward pass. The retrieval is therefore query-conditioned from the first evolution round.  After each latent memory updating, the relevance scores are recomputed from the evolved LMTs, allowing historical access to adapt progressively to what has already been internalized and what remains unresolved.

\noindent\textbf{Group-Wise Evidence Injection and Latent Memory Updating.} Given the candidate evidence \(\mathbf{E}_{g}^{(r)}\) retrieved above, we instantiate an optimizable copy \(\widetilde{\mathbf{E}}_{g}^{(r)}\). Let \(\widehat{\mathbf{E}}_{g}^{(r-1)}\) denote the visual evidence pool accepted up to the previous iteration. To preserve the association between each LMT group and its retrieved evidence, we insert
\(\widehat{\mathbf{E}}_{g}^{(r-1)}\) and
\(\widetilde{\mathbf{E}}_{g}^{(r)}\)
immediately after the corresponding LMT group
\(\widehat{\mathbf{Z}}_{g}^{(r-1)}\)
in the candidate input \(\mathbf{X}_{\mathrm{cand}}^{(r)}\).
Rather than modifying the preceding LMT activations through backward attention within the same forward pass, the retrieved evidence conditions the reward signal, whose gradient subsequently updates the optimizable LMT embeddings.
In this way, the retrieved evidence is progressively internalized into the latent memory across optimization iterations.

All injected evidence representations and LMT embeddings are jointly optimized under the progressive confidence objective (\S\ref{sec:progressive_confidence_optimization}). Let \(\mathcal{R}_{\mathrm{cand}}^{(r)}\) denote the reward obtained with the current proposal and \(\widehat{\mathcal{R}}^{(r-1)}\) the reward preserved from the previous iteration. The newly retrieved evidence is admitted into the visual evidence pool only when the candidate state yields a reward improvement:
\begin{equation}
\begin{aligned}
\widehat{\mathbf{E}}_{g}^{(r)}
&=
\begin{cases}
\operatorname{Merge}\!\left(
    \widehat{\mathbf{E}}_{g}^{(r-1)},
    \widetilde{\mathbf{E}}_{g}^{(r)}
\right),
&
\text{if }\;
\mathcal{R}_{\mathrm{cand}}^{(r)}
>
\widehat{\mathcal{R}}^{(r-1)}, \\[1mm]
\widehat{\mathbf{E}}_{g}^{(r-1)},
&
\text{otherwise},
\end{cases}
\end{aligned}
\label{eq:reward_gated_evidence_accumulation}
\end{equation}
where \(\operatorname{Merge}(\cdot,\cdot)\) appends the newly selected visual evidence to the previously retained pool while removing duplicated entries.
This reward gate controls only the accumulation of retrieved evidence: informative evidence is preserved across evolution iterations, whereas evidence that fails to improve the candidate reward is discarded. The updated LMTs subsequently guide the next retrieval iteration. The detailed optimization of the latent memory tokens is presented in
\S\ref{sec:progressive_confidence_optimization}.

\subsection{Progressive Confidence-guided Latent Memory Optimization}
\label{sec:progressive_confidence_optimization}

\noindent\textbf{Hierarchical Progression Reward Objective.}
The three LMT groups integrate query-relevant evidence from recent observations to the complete historical memory. However, broader retrieval may also introduce irrelevant evidence and destabilize latent evolution. We therefore exploit the frozen MLLM's predictive uncertainty as intrinsic feedback, encouraging the latent memory to become increasingly confident as its accessible historical scope expands.

At evolution iteration \(r\),
let \(b_g\) denote the terminal position of the LMT--evidence block associated with group \(g\).
For an optimizable latent state \(\mathbf{U}\), we obtain the predictive distribution
\(\mathbf{p}_g(\mathbf{U})\)
and compute its normalized top-\(\delta\) entropy:
\begin{equation}
\mathcal{E}_g(\mathbf{U})
=
-\frac{1}{\log\delta}
\sum_{j\in\Omega_g(\mathbf{U})}
\bar{p}_{g,j}(\mathbf{U})
\log\bar{p}_{g,j}(\mathbf{U}),
\qquad
\bar{p}_{g,j}(\mathbf{U})
=
\frac{p_{g,j}(\mathbf{U})}
{\sum_{v\in\Omega_g(\mathbf{U})}p_{g,v}(\mathbf{U})},
\label{eq:groupwise_latent_entropy}
\end{equation}
where
\(\Omega_g(\mathbf{U})
=
\operatorname{TopK}_{\delta}(\mathbf{p}_g(\mathbf{U}))\) denotes the set of the $\delta$ tokens with the highest probabilities.
Since each group-wise distribution is evaluated after its corresponding retrieved evidence, the resulting entropy measures evidence-aware predictive confidence.
A lower entropy indicates higher predictive confidence.
Since the three groups incorporate increasingly comprehensive historical evidence, the desired progression is
\(\mathcal{E}_s(\mathbf{U})>\mathcal{E}_m(\mathbf{U})>\mathcal{E}_l(\mathbf{U})\).
To encourage both absolute confidence and consistent hierarchical progression, we define the hierarchical progression reward as:
\begin{equation}
\begin{aligned}
\mathcal{R}_{\mathrm{prog}}(\mathbf{U})
=&\
\underbrace{
\sum\nolimits_{g\in\mathcal{G}}
\left(
1-\mathcal{E}_g(\mathbf{U})
\right)
}_{\text{absolute confidence}}
+
\underbrace{
\left(
\left[
\mathcal{E}_s(\mathbf{U})
-
\mathcal{E}_m(\mathbf{U})
\right]_{+}
+
\left[
\mathcal{E}_m(\mathbf{U})
-
\mathcal{E}_l(\mathbf{U})
\right]_{+}
\right)
}_{\text{hierarchical progression}},
\end{aligned}
\label{eq:hierarchical_progression_reward}
\end{equation}
where $[x]_{+}=\max(0,x)$. This objective encourages the
latent memory to become both absolutely and progressively more confident as more historical evidence is incorporated.

\noindent\textbf{Confidence-guided Latent Memory Optimization.}
At evolution iteration $r$, the latent memory tokens and retrieved evidence jointly form the optimizable latent state $\mathbf{U}^{(r,0)}$.
We first perturb these continuous representations by sampling a Gaussian distribution
$\boldsymbol{\xi}\sim
\mathcal{N}(\mathbf{0},\sigma^2\mathbf{I})$
and forming $\mathbf{U}'=\mathbf{U}+\boldsymbol{\xi}$, where $\sigma$ controls the magnitude of exploration.
This induces the Gaussian policy
$\pi_{\sigma}(\mathbf{U}'\mid\mathbf{U})
=
\mathcal{N}(\mathbf{U},\sigma^2\mathbf{I})$.
Starting from $\mathbf{U}^{(r,0)}$, we adopt a REINFORCE-based direct policy-gradient method~\cite{williams1992simple} and iteratively update the latent state as:
\begin{equation}
    \mathbf{U}
    \leftarrow
    \mathbf{U}
    +
    \eta
    \widehat{\nabla}_{\mathbf{U}}
    \mathcal{J}(\mathbf{U}),
    \label{eq:test_time_latent_update}
\end{equation}
where $\eta$ is the learning rate and
$\mathcal{J}(\mathbf{U})
=
\mathbb{E}_{\mathbf{U}'\sim
\pi_{\sigma}(\cdot\mid\mathbf{U})}
[\mathcal{R}_{\mathrm{prog}}(\mathbf{U}')]$
denotes the expected progression reward.
Following the policy-gradient theorem, its gradient can be expressed as:
\begin{equation}
\begin{aligned}
    \nabla_{\mathbf{U}}\mathcal{J}(\mathbf{U})
    &=
    \mathbb{E}_{\mathbf{U}'\sim
    \pi_{\sigma}(\cdot\mid\mathbf{U})}
    \left[
        \mathcal{R}_{\mathrm{prog}}(\mathbf{U}')
        \nabla_{\mathbf{U}}
        \log
        \pi_{\sigma}(\mathbf{U}'\mid\mathbf{U})
    \right]=
    \mathbb{E}
    \left[
        \mathcal{R}_{\mathrm{prog}}(\mathbf{U}')
        \frac{\boldsymbol{\xi}}{\sigma^2}
    \right].
\end{aligned}
\label{eq:reinforce_latent_gradient}
\end{equation}
Since \(\mathcal{R}_{\mathrm{prog}}(\mathbf{U}')\) is evaluated on the candidate state containing group-wise retrieved evidence, its gradient provides evidence-conditioned updates to the LMT embeddings.
This jointly optimizes the LMTs and retrieved evidence, progressively internalizing task-relevant history without external supervision or model-parameter updates.

After \(R\) evolution iterations, all injected evidence copies are removed, and the final answer is decoded from
\(\left[\mathcal{M};q;\widehat{\mathbf{Z}}^{(R)}\right]\),
allowing the compact latent memory to guide more confident reasoning for final Streaming question answering.

\begin{table}[t]
\centering
\caption{
\textbf{Comparison with state-of-the-art methods on OVO-Bench~\cite{ovobench}}.
Best results among open-source models are in \textbf{bold}, and the best results
among training-free methods are \underline{underlined}.
$^\dagger$ indicates the reproduced results.
}
\resizebox{1.0\linewidth}{!}{
\setlength{\tabcolsep}{3.0pt}
\renewcommand{\arraystretch}{1.05}
\label{tab:ovo_bench}
\begin{tabular}{lcc|cccccc|c|ccc|c|c}
\thickhline
\multirow{2}{*}{\textbf{Method}}
& \multirow{2}{*}{\textbf{Size}}
& \multirow{2}{*}{\textbf{Frames}}
& \multicolumn{7}{c|}{\textbf{Real-Time Visual Perception}}
& \multicolumn{4}{c|}{\textbf{Backward Tracing}}
& \multirow{2}{*}{\textbf{Overall}}
\\
\cline{4-10}
\cline{11-14}
& & &
\textbf{OCR} &
\textbf{ACR} &
\textbf{ATR} &
\textbf{STU} &
\textbf{FPD} &
\textbf{OJR} &
\textbf{Avg.} &
\textbf{EPM} &
\textbf{ASI} &
\textbf{HLD} &
\textbf{Avg.} &
\\
\hline

\rowcolor{groupgray}
\multicolumn{15}{c}{\textit{Proprietary Models}} \\
\hline

Gemini 1.5 Pro~\cite{team2024gemini}
& -- & 1 fps
& 85.9 & 67.0 & 79.3 & 58.4 & 63.4 & 62.0 & 69.3
& 58.6 & 76.4 & 52.6 & 62.5
& 65.9
\\

GPT-4o~\cite{hurst2024gpt}
& -- & 64
& 69.8 & 64.2 & 71.6 & 51.1 & 70.3 & 59.8 & 64.5
& 57.9 & 75.7 & 48.7 & 60.8
& 62.6
\\

\hline

\rowcolor{groupgray}
\multicolumn{15}{c}{\textit{Open-source Offline MLLMs}} \\
\hline

LLaVA-Video~\cite{zhang2024llava}
& 7B & 64
& 69.8 & 59.6 & 66.4 & 50.6 & 72.3 & 61.4 & 63.3
& 51.2 & 64.2 & 9.7 & 41.7
& 52.5
\\

Qwen2-VL~\cite{wang2024qwen2}
& 7B & 64
& 69.1 & 53.2 & 63.8 & 50.6 & 66.3 & 60.9 & 60.7
& 44.4 & \textbf{66.9} & 34.4 & 48.6
& 54.6
\\

InternVL2~\cite{chen2024far}
& 8B & 64
& 68.5 & 58.7 & 69.0 & 44.9 & 67.3 & 56.0 & 60.7
& 43.1 & 61.5 & 27.4 & 44.0
& 52.4
\\

LongVU~\cite{shen2024longvu}
& 7B & 1 fps
& 55.7 & 49.5 & 59.5 & 48.3 & 68.3 & 63.0 & 57.4
& 43.1 & 66.2 & 9.1 & 39.5
& 48.5
\\

\hline

\rowcolor{groupgray}
\multicolumn{15}{c}{\textit{Open-source Online MLLMs (Training-Based)}} \\
\hline

VideoLLM-Online~\cite{chen2024videollm}
{\scriptsize\color{gray}[CVPR 2024]}
& 8B & 2 fps
& 8.1 & 23.9 & 12.1 & 14.0 & 45.5 & 21.2 & 20.8
& 22.2 & 18.8 & 12.2 & 17.7
& 19.3
\\

Dispider~\cite{qian2025dispider}
{\scriptsize\color{gray}[CVPR 2025]}
& 7B & 1 fps
& 57.7 & 49.5 & 62.1 & 44.9 & 61.4 & 51.6 & 54.6
& 48.5 & 55.4 & 4.3 & 36.1
& 45.3
\\

Flash-VStream~\cite{zhang2025flash}
{\scriptsize\color{gray}[ICCV 2025]}
& 7B & 1 fps
& 25.5 & 32.1 & 29.3 & 33.7 & 29.7 & 28.8 & 29.9
& 36.4 & 33.8 & 5.9 & 25.4
& 27.6
\\

ViSpeak~\cite{fu2025vispeak}
{\scriptsize\color{gray}[ICCV 2025]}
& 7B & 1 fps
& 75.2 & 58.7 & 71.6 & 51.1 & 74.3 & 66.9 & 66.3
& \textbf{59.9} & 48.7 & 64.0 & 57.5
& 61.9
\\

TimeChat-Online~\cite{timechat}
{\scriptsize\color{gray}[ACM MM 2025]}
& 7B & 1 fps
& 75.2 & 46.8 & 70.7 & 47.8 & 69.3 & 61.4 & 61.9
& 55.9 & 59.5 & 9.7 & 41.7
& 51.8
\\

StreamForest~\cite{zeng2026streamforest}
{\scriptsize\color{gray}[NeurIPS 2025]}
& 7B & 1 fps
& 68.5 & 53.2 & 71.6 & 47.8 & 65.4 & 60.9 & 61.2
& 58.9 & 64.9 & 32.3 & 52.0
& 56.6
\\

Streamo~\cite{xia2026streaming}
{\scriptsize\color{gray}[CVPR 2026]}
& 7B & 1 fps
& 79.2 & 57.8 & \textbf{75.0} & 49.4 & 64.4 & \textbf{70.1} & 66.0
& 54.6 & 52.0 & 31.7 & 46.1
& 56.1
\\

ThinkStream~\cite{liu2026thinking}
{\scriptsize\color{gray}[ECCV 2026]}
& 3B & 1 fps
& 85.2 & 64.2 & 69.8 & 49.4 & 69.3 & 64.1 & 67.0
& 53.9 & 59.5 & 43.6 & 52.3
& 59.7
\\

\hline

\rowcolor{groupgray}
\multicolumn{15}{c}{\textit{Open-source Online MLLMs (Training-Free)}} \\
\hline

Qwen2.5-VL-3B$^\dagger$~\cite{qwen25}
& 3B & 1 fps
& 77.2 & 52.3 & 69.0 & 41.0 & 67.3 & 60.9 & 61.3
& 49.8 & 53.4 & 26.3 & 43.2
& 52.2
\\

+ FluxMem
& 3B & 1 fps
& 83.2 & 56.9 & 67.2 & 47.8 & 68.3 & 63.6 & 64.5
& 47.5 & 54.7 & 24.2 & 42.1
& 53.3
\\

\rowcolor{oursblue}
\textbf{+ \ourmethod (Ours)}
& 3B & 1 fps
& 84.6 & 57.8 & 70.7 & 46.1 & 71.3 & 64.1 & 65.8
& 51.2 & 53.4 & 51.6 & 52.1
& 59.0 {\color{gaincolor}(+6.8)}
\\

\hline

Qwen2.5-VL-7B$^\dagger$~\cite{qwen25}
& 7B & 1 fps
& 79.2 & 53.2 & 67.2 & 51.7 & 71.3 & 57.1 & 63.3
& 51.5 & 58.8 & 23.7 & 44.7
& 54.0
\\

+ QueryStream~\cite{zhang2026querystream}
{\scriptsize\color{gray}[ICLR 2026]}
& 7B & 1 fps
& 75.2 & 49.5 & 69.8 & 50.0 & 71.3 & 62.5 & 63.1
& \underline{56.9} & \underline{65.5} & 12.4 & 44.9
& 54.0
\\

+ FluxMem~\cite{fluxmem}
{\scriptsize\color{gray}[CVPR 2026]}
& 7B & 1 fps
& 81.2 & 59.6 & 70.7 & \textbf{\underline{53.4}} & 75.2 & 63.0 & 67.2
& 50.2 & 62.8 & 26.9 & 46.6
& 56.9
\\

+ OASIS~\cite{oasis}
{\scriptsize\color{gray}[CVPR 2026]}
& 7B & --
& 85.2 & \textbf{\underline{72.5}} & 66.4 & 52.3 & 67.3 & 64.7 & 67.3
& 51.9 & 58.8 & 48.9 & 52.6
& 60.0
\\

\rowcolor{oursblue}
\textbf{+ \ourmethod (Ours)}
& 7B & 1 fps
& \textbf{\underline{85.9}}
& 61.5
& \underline{73.3}
& 46.6
& \textbf{\underline{78.2}}
& \underline{65.2}
& \textbf{\underline{68.5}}
& 50.8
& 60.8
& \textbf{\underline{68.3}}
& \textbf{\underline{60.0}}
& \textbf{\underline{64.2}} {\color{gaincolor}(+10.2)}
\\
\hline
\end{tabular}
}
\end{table}
\section{Experiment}
\label{sec::experi}
\subsection{Experiment Setup}
\noindent\textbf{Benchmarks.}
We evaluate \ourmethod on two streaming video benchmarks and three offline long-video benchmarks. OVO-Bench~\cite{ovobench} evaluates timestamp-aware streaming video understanding, covering historical retrieval, real-time perception, and proactive response. StreamingBench~\cite{streamingbench} evaluates real-time visual, omni-source, and contextual understanding over continuous video streams. 
For offline evaluation, Video-MME~\cite{videomme}, MLVU~\cite{mlvu}, and LongVideoBench~\cite{longvideobench} collectively assess perception, retrieval, and reasoning over long videos across varying durations and levels of temporal granularity.

\noindent\textbf{Implementation Details.}
we build our \ourmethod on Qwen2.5-VL-3B and 7B backbones~\cite{qwen25}. For online benchmarks, videos are sampled at 1 fps with at most 256 frames. The short- and mid-term memory capacities are set to 8 and 64 frames, respectively, while earlier observations are consolidated into long-term memory under a global budget of 2048 visual tokens.
The number of latent memory tokens $K$ for each group is set to $2$,  with $B=8$ visual candidate patches injected at each iteration.
Unless otherwise specified, we perform $R=4$ evolution iterations with a learning rate of $\eta=1\times10^{-3}$ and a Gaussian perturbation scale of $\sigma=10\%$.
For offline benchmarks, videos are sampled at the same frame rate with $64$ visual tokens per frame and at most $1,024$ frames. 
All experiments are conducted on eight NVIDIA H20 GPUs. Further implementation details are provided in the Appendix.

\subsection{Comparison with State-of-the-Arts}
\noindent\textbf{Results on Streaming Video Benchmarks.} As shown in Tables~\ref{tab:ovo_bench} and \ref{tab:streaming_offline}, \ourmethod consistently improves the baseline Qwen2.5-VL~\cite{qwen25} across both streaming benchmarks while keeping the backbone frozen. On OVO-Bench~\cite{ovobench}, our \ourmethod (7B) improves Real-Time Visual Perception from 63.3\% to \textbf{68.5\%} and Backward Tracing from 44.7\% to \textbf{60.0\%}, achieving the best overall score of \textbf{64.2\%} (\textbf{+10.2\%}) among open-source methods. our \ourmethod (3B) similarly improves the overall score from 52.2\% to \textbf{59.0\%} (\textbf{+6.8\%}).
On StreamingBench~\cite{streamingbench}, \ourmethod reaches \textbf{76.9\%} (\textbf{+3.0\%}), outperforming all compared training-free methods. These results demonstrate the consistent effectiveness of progressive latent memory across diverse streaming video understanding tasks.

\begin{table}[t]
\centering
\caption{
\textbf{Comparison with state-of-the-art methods on StreamingBench~\cite{streamingbench} and offline video benchmarks~\cite{videomme,mlvu,longvideobench}}.
Best results among open-source models are in \textbf{bold}, and the best results
among training-free methods are \underline{underlined}.
$^\dagger$ indicates the reproduced results.
}
\label{tab:streaming_offline}
\small
\resizebox{1.0\linewidth}{!}{
\setlength{\tabcolsep}{2.0pt}
\renewcommand{\arraystretch}{1.05}
\begin{tabular}{lcc|c|cccc|c|c}
\thickhline

\multirow{3}{*}{\textbf{Method}}
& \multirow{3}{*}{\textbf{Size}}
& \multirow{3}{*}{\textbf{Frames}}
& \multicolumn{1}{c|}{\textbf{Online Video}}
& \multicolumn{6}{c}{\textbf{Offline Video}}
\\

\cline{4-4}
\cline{5-10}

& & &
\textbf{StreamingBench}
&
\multicolumn{4}{c|}{\textbf{VideoMME}}
&
\textbf{MLVU}
&
\textbf{LongVideoBench}
\\

\cline{4-4}
\cline{5-8}

& & &
\textbf{Real-Time}
&
\textbf{Short}
&
\textbf{Medium}
&
\textbf{Long}
&
\textbf{All}
&
\textbf{M-Avg}
&
\textbf{Val}
\\

\hline

\rowcolor{groupgray}
\multicolumn{10}{c}{\textit{Open-source Offline MLLMs}} \\
\hline

LLaVA-Video~\cite{zhang2024llava}
& 7B & 1 fps
& --
& -- & -- & -- & 63.3
& 70.8
& --
\\

LLaVA-OneVision~\cite{Llavaonevision}
& 7B & 32
& 71.1
& 70.1 & 56.4 & 48.8 & 58.4
& 64.7
& 56.5
\\

InternVL2.5~\cite{chen2024far}
& 8B & 64
& --
& -- & -- & -- & 64.2
& 68.9
& 60.0
\\
LongVU~\cite{shen2024longvu}
& 7B & 1 fps
& --
& -- & -- & -- & 60.6
& 65.4
& --
\\
\hline

\rowcolor{groupgray}
\multicolumn{10}{c}{\textit{Open-source Online MLLMs (Training-Based)}} \\
\hline

VideoLLM-Online~\cite{chen2024videollm}
{\scriptsize\color{gray}[CVPR 2024]}
& 8B & 2 fps
& 36.0
& -- & -- & -- & --
& --
& --
\\

Dispider~\cite{qian2025dispider}
{\scriptsize\color{gray}[CVPR 2025]}
& 7B & 1 fps
& 67.6
& -- & 53.7 & 49.7 & 57.2
& 61.7
& --
\\

Flash-VStream~\cite{zhang2025flash}
{\scriptsize\color{gray}[ICCV 2025]}
& 7B & 1 fps
& 23.2
& 72.0 & 61.1 & 50.3 & 61.2
& --
& --
\\

TimeChat-Online~\cite{timechat}
{\scriptsize\color{gray}[ACM MM 2025]}
& 7B & 1 fps
& 75.3
& -- & -- & 52.4 & 63.3
& 65.4
& 57.7
\\

StreamForest~\cite{zeng2026streamforest}
{\scriptsize\color{gray}[NeurIPS 2025]}
& 7B & 1 fps
& \textbf{77.3}
& -- & -- & -- & 61.9
& 69.6
& --
\\
ThinkStream~\cite{liu2026thinking}
{\scriptsize\color{gray}[ECCV 2026]}
& 3B & 1 fps
& 75.0
& -- & -- & -- & 61.9
& --
& 56.4
\\

\hline

\rowcolor{groupgray}
\multicolumn{10}{c}{\textit{Open-source Online MLLMs (Training-Free)}} \\
\hline

ReKV~\cite{di2025streaming}
{\scriptsize\color{gray}[ICLR 2025]}
& 7B & 0.5 fps
& 69.1
& -- & -- & -- & --
& 68.5
& --
\\

StreamChat~\cite{xiong2025streaming}
{\scriptsize\color{gray}[ICLR 2025]}
& 8B & 1 fps
& 64.7
& -- & -- & -- & --
& --
& --
\\

\hline

Qwen2.5-VL$^\dagger$~\cite{qwen25}
& 7B & 1 fps
& 73.9
& 73.8
& 62.4
& 53.8
& 63.3
& 67.9
& 60.7
\\
+  QueryStream~\cite{zhang2026querystream}
{\scriptsize\color{gray}[ICLR 2026]}
& 7B & 1 fps
 &75.3 &--  &--  &49.8  & 63.2  &--  &58.0
\\
+ FluxMem~\cite{fluxmem}
{\scriptsize\color{gray}[CVPR 2026]}
& 7B & 1 fps
 &76.4 &76.9  &65.1  &54.0  & 65.3  &73.1  &61.1 
\\
+ OASIS~\cite{oasis}
{\scriptsize\color{gray}[CVPR 2026]}
& 7B & --
 &70.6 &--  &--  &--  & --  &--  &-- 
\\

\rowcolor{oursblue}
\textbf{+ \ourmethod (Ours)}
& 7B & 1 fps
& \underline{76.9}
& \textbf{\underline{77.2}}
& \textbf{\underline{67.4}}
& \textbf{\underline{55.1}}
& \textbf{\underline{66.6}}
& \textbf{\underline{74.0}}
& \textbf{\underline{62.1}}
\\
\hline
\end{tabular}}
\vspace{-6pt}
\end{table}

\begin{table*}[t]
\centering

\begin{minipage}[t]{0.42\linewidth}
\vspace{0pt}
\centering
\captionof{table}{
\textbf{Impacts of core components} on OVO-Bench~\cite{ovobench} and VideoMME~\cite{videomme}.
}
\label{tab:ablation_components}
 \vspace{-6pt}
\small
\resizebox{\linewidth}{!}{
\setlength{\tabcolsep}{3.3pt}
\renewcommand{\arraystretch}{1.05}
\begin{tabular}{ccc||cc}
\thickhline
\rowcolor{groupgray}
\textbf{HSM} & \textbf{PMO} & \textbf{HME}
& \textbf{OVO-Bench} & \textbf{VideoMME} \\
\hline\hline
            &            &            & 54.0 & 63.3 \\
\checkmark  &            &            & 58.1 & 65.1 \\
\checkmark  & \checkmark &            & 62.4 & 66.1 \\
\arrayrulecolor{gray}
\hdashline
\arrayrulecolor{black}
\checkmark  & \checkmark & \checkmark
& \textbf{64.2} & \textbf{66.6} \\
\hline
\end{tabular}
}
\end{minipage}
\hfill
\begin{minipage}[t]{0.57\linewidth}
\vspace{0pt}
\centering
\captionof{table}{
\textbf{Ablation of retrieve-and-internalize mechanism}
on OVO-Bench~\cite{ovobench} and VideoMME~\cite{videomme}.
}
\label{tab:ablation_internalization}
 \vspace{-6pt}
\small
\resizebox{\linewidth}{!}{
\setlength{\tabcolsep}{2.6pt}
\renewcommand{\arraystretch}{1.05}
\begin{tabular}{l||cc}
\thickhline
\rowcolor{groupgray}
\textbf{Method}
& \textbf{OVO-Bench}
& \textbf{VideoMME} \\
\hline\hline
\textsc{Baseline}
& 56.9 & 65.4 \\
+ Retrieved Visual Evidence
& 59.7 & 65.6 \\
+ Initial Latent Memory Tokens
& 56.5 & 65.2 \\
\arrayrulecolor{gray}
\hdashline
\arrayrulecolor{black}
+ Evolved Latent Memory Tokens (\textbf{Ours})
& \textbf{64.2}
& \textbf{66.6} \\
\hline
\end{tabular}
}
\end{minipage}

\end{table*}
\noindent\textbf{Results on Offline Video Benchmarks.}   
Though designed for streaming video understanding, \ourmethod generalizes effectively to offline long-video understanding under a bounded memory budget. As shown in Table~\ref{tab:streaming_offline}, \ourmethod achieves \textbf{66.6\%} on VideoMME~\cite{videomme}, \textbf{74.0\%}  on MLVU~\cite{mlvu}, and \textbf{62.1\%}  on LongVideoBench~\cite{longvideobench}, outperforming the Qwen2.5-VL-7B~\cite{qwen25} baseline by \textbf{3.3\%} , \textbf{6.1\%}, and \textbf{1.4\%}  points, respectively. These results surpass all compared training-free and training-based methods. On VideoMME, \ourmethod further improves the short-, medium-, and long-video scores from 73.8\%, 62.4\%, and 53.8\%  to\textbf{ 77.2\%}, \textbf{67.4\%}, and \textbf{55.1\%}, respectively. The consistent gains across different video durations demonstrate that hierarchical streaming memory and progressive latent memory evolution effectively preserve and internalize task-relevant historical evidence beyond the online setting.

\subsection{Ablation Study}
\noindent\textbf{Key Component Analysis.}
Table~\ref{tab:ablation_components} first evaluates the contribution of three core components. Query-agnostic Hierarchical Streaming Memory (HSM) improves OVO-Bench~\cite{ovobench} and VideoMME~\cite{videomme} from 54.0\%/63.3\% to \textbf{58.1\%}/\textbf{65.1\%}, validating the benefit of hierarchical memory consolidation. Adding Progressive Confidence-guided Latent Memory Optimization (PMO) further raises the results to \textbf{62.4\%}/\textbf{66.1\%}. Although visual tokens are neither retrieved nor injected in this variant, the gains demonstrate that confidence-guided optimization alone can effectively refine the latent memory tokens (LMTs). Finally, Hierarchical Latent Memory Evolution (HME) achieves the best performance of \textbf{64.2\%}/\textbf{66.6\%} by retrieving visual evidence from expanding memory scopes and injecting it into LMT evolution, validating the effectiveness of our progressive latent working memory. Overall, the full model outperforms the baseline by \textbf{10.2\%} and \textbf{3.3\%}, confirming the complementarity of the three components.

\noindent\textbf{Effectiveness of Retrieve-and-internalize Latent Memory.} 
To distinguish latent internalization from simple context augmentation, Table~\ref{tab:ablation_internalization} contrasts four ways of exploiting historical information.
 The baseline, conditioned on only the hierarchical memory and query (\ie, FluxMem~\cite{fluxmem}), obtains 56.9\%/65.4\% on OVO-Bench~\cite{ovobench}/VideoMME~\cite{videomme}. 
 Directly appending retrieved visual evidence following~\cite{zhang2026querystream,oasis} improves the results to 59.7/65.6 on OVO-Bench/VideoMME, showing that relevant evidence is beneficial but remains insufficient when used merely as external context. Introducing unoptimized latent memory tokens (LMTs) instead yields 56.5\%/65.2\%, indicating that the gains cannot be attributed to simply adding latent tokens. In contrast, the evolved LMTs achieve 64.2\%/66.6\%, outperforming direct evidence injection by \textbf{4.5\%}/\textbf{1.0\%} and initial LMTs by \textbf{7.7\%}/\textbf{1.4\%} points. Notably, the retrieved visual tokens are removed before final decoding. These results demonstrate that the retrieved evidence is effectively internalized into a compact latent working memory for streaming reasoning.

\begin{wraptable}{r}{0.58\linewidth}
\vspace{-12pt}
\centering
\caption{
\textbf{Ablation of progressive confidence-guided optimization}
on OVO-Bench~\cite{ovobench} and VideoMME~\cite{videomme}.}
\label{tab:ablation_reward}
\small
\vspace{-6pt}
\resizebox{1.0\linewidth}{!}{
\setlength{\tabcolsep}{3pt}
\renewcommand{\arraystretch}{1.05}
\begin{tabular}{l||cc}
\thickhline
\rowcolor{groupgray}
\textbf{Confidence Objective}
& \textbf{OVO-Bench}
& \textbf{VideoMME} \\
\hline\hline
w/o Latent Memory Optimization
& 58.1 & 65.1 \\
Absolute Confidence Optimization \textit{only}
& 62.5 & 65.5 \\
\arrayrulecolor{gray}
\hdashline
\arrayrulecolor{black}
Hierarchical  Progression Reward (\textbf{Ours})
& \textbf{64.2}
& \textbf{66.6} \\

\hline
\end{tabular}}
\vspace{-13pt}
\hspace{-2ex}
\end{wraptable}

\noindent\textbf{Effect of Progressive Confidence-guided Latent Memory Optimization.} 
To study the effectiveness of our hierarchical progression reward (\cf~Eq.~\ref{eq:hierarchical_progression_reward}), Table~\ref{tab:ablation_reward} compares it against no latent memory optimization and absolute confidence optimization alone. Without latent memory optimization, the model obtains 58.1\%/65.1\% on OVO-Bench~\cite{ovobench}/VideoMME~\cite{videomme}. Optimizing only the absolute confidence term, $\sum_{g\in\mathcal{G}}(1-\mathcal{E}_g)$, improves the results to 62.5\%/65.5\%, showing that predictive confidence provides effective self-supervision for refining latent memory tokens and retrieved evidence without annotations. However, this objective encourages each group to become confident independently, without considering whether confidence increases consistently with the expanding memory receptive fields. By additionally encouraging the entropy ordering $\mathcal{E}_s>\mathcal{E}_m>\mathcal{E}_l$, our hierarchical progression reward achieves \textbf{64.2\%}/\textbf{66.6\%}, outperforming absolute confidence optimization by \textbf{1.7\%}/\textbf{1.1\%}. These results confirm that progressive confidence-guided latent memory optimization provides stronger test-time supervision for latent memory evolution.

\begin{figure}[!t]
    \centering
    \includegraphics[width=\linewidth]{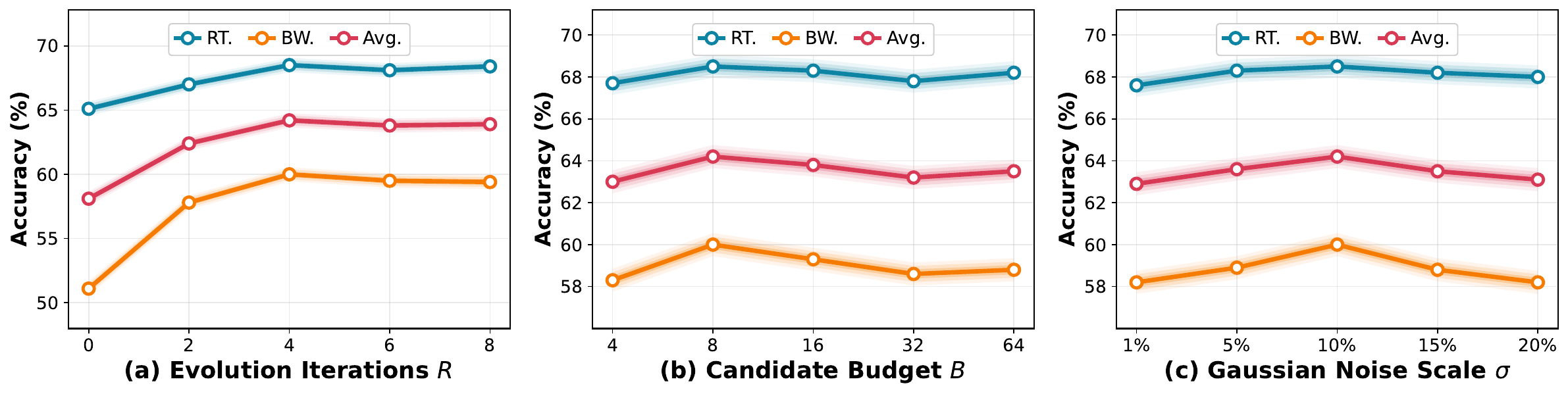}
    \vspace{-15pt}
    \caption{\textbf{Hyperparameter analysis on OVO-Bench~\cite{ovobench}}. (a) Effect of the number of evolution iterations $R$. (b) Effect of the candidate budget $B$. (c) Effect of Gaussian noise scale $\sigma$. “RT” / “BW” denote Real-Time Visual Perception and Backward Tracing; “Avg.” is the mean of RT and BW.}
    \label{fig:params}
    \vspace{-1.4em}
\end{figure}

\noindent\textbf{Evolution Iteration Number $R$.}
To evaluate the effect of the evolution iteration number $R$ on streaming video understanding, we conduct experiments with different values of $R$, as shown in Fig.~\ref{fig:params}~(a). The results indicate that increasing $R$ from 0 to 4 consistently improves performance, raising the average accuracy from 58.1\% to \textbf{64.2\%}, which demonstrates that iterative retrieval--update optimization effectively enhances latent memory evolution. Although minor fluctuations occur with further iterations, the model maintains high accuracy and shows no additional improvement. We therefore fix $R=4$ to balance performance and efficiency.

\noindent\textbf{Candidate Budget $B$.}
As shown in Fig.~\ref{fig:params}~(b), we study the effect of varying the candidate budget (\ie, injected visual evidence each iteration) on streaming video understanding. As $B$ increases from 4 to 8, the average accuracy improves from 63.0\% to \textbf{64.2\%}; however, further increasing the budget leads to performance degradation. This suggests that a limited set of relevant visual evidence is sufficient to effectively guide latent memory evolution, whereas excessive candidates introduce redundancy that hinders optimization.

\noindent\textbf{Gaussian noise scale $\sigma$.}
We further investigate how the Gaussian perturbation scale $\sigma$ governs the dynamics of latent memory optimization. As illustrated in Fig.~\ref{fig:params}~(c), increasing $\sigma$ from 1\% to 10\% enhances latent exploration, enabling the model to explore a broader range of optimization trajectories and discover higher-confidence memory states. However, when $\sigma$ becomes excessively large, the stronger perturbations destabilize the optimization process, resulting in performance degradation.
Therefore, we set $\sigma=10\%$ for a well-calibrated level of perturbation.

\begin{wraptable}{r}{0.58\linewidth}
\vspace{-12pt}
    \centering
    \caption{
        \textbf{Efficiency comparison} on OVO-Bench~\cite{ovobench}
        with Qwen2.5-VL-7B as the backbone.
    }
    \label{tab:efficiency_comparison}
    \vspace{-8pt}
    \small
    \resizebox{1.0\linewidth}{!}{
    \setlength{\tabcolsep}{5pt}
    \renewcommand{\arraystretch}{1.08}
    \begin{tabular}{l||cccc}
       \thickhline
\rowcolor{groupgray}
        \textbf{Method}
        & \textbf{Peak Mem$\downarrow$}
        & \textbf{TTFT$\downarrow$}
        & \textbf{TPOT$\downarrow$}
        & \textbf{Perf.$\uparrow$} \\
       \hline\hline

        Qwen2.5-VL-7B
        & 30.80GB & 7.63s & 6.45ms & 54.0 \\

        + \ourmethod \textbf{(Ours)}
        & 21.97GB
        & 8.41s
        & 3.16ms
        &  64.2 \\
        \hline
    \end{tabular}}
\vspace{-12pt}
\hspace{-2ex}
\end{wraptable}

\noindent\textbf{Efficiency Analysis.}
We further evaluate the efficiency of \ourmethod under the same Qwen2.5-VL-7B backbone on OVO-Bench~\cite{ovobench}. As shown in Table~\ref{tab:efficiency_comparison}, \ourmethod reduces peak memory from 30.80 GB to 21.97 GB while improving performance from 54.0 to 64.2 (\textbf{+18.9\%}). It also decreases time per output token (TPOT) from 6.45 ms to 3.16 ms, yielding a \textbf{51.0\% reduction} in per-token decoding latency. This improvement demonstrates that the compact hierarchical memory effectively limits the accumulated visual context and reduces autoregressive decoding overhead. Meanwhile, the time to first token (TTFT) increases from 7.63 s to 8.41 s due to the additional latent memory evolution performed before answer generation. Overall, \ourmethod provide a favorable accuracy--efficiency balance for streaming video understanding.




\section{Conclusion}


In this work, we introduced \ourmethod, a progressive latent working memory framework for streaming video understanding. Moving beyond conventional store-and-retrieve memory, \ourmethod progressively internalizes task-relevant historical evidence into a compact, query-conditioned latent memory that continuously guides streaming reasoning. It integrates Query-Agnostic Hierarchical Streaming Memory, Hierarchical Latent Memory Evolution, and Progressive Confidence-guided Latent Memory Optimization, while keeping the underlying MLLM fully frozen. Extensive experiments across streaming and offline long-video benchmarks demonstrate the effectiveness of \ourmethod. These results establish retrieve-and-internalize memory as a promising direction for bridging external memory with latent reasoning in streaming Video-LLMs.

\bibliography{iclr2027_conference}
\bibliographystyle{iclr2027_conference}


\end{document}